\documentclass[acmsmall,screen,authorversion,nonacm]{acmart}

\usepackage{fancyhdr}
\AtBeginDocument{%
    \addtolength{\footskip}{2.0\baselineskip}%
    \fancyfoot[L]{\textit{\textbf{Author Preprint.}}}%
}

\AtBeginDocument{%
  }

\usepackage{geometry}
\usepackage{array}
\usepackage{enumitem}
\usepackage{listings}
\usepackage{color}
\usepackage{xcolor}
\usepackage{hyperref}
\usepackage{xtab}  % Use xtab for multipage sideways tables
\usepackage{pdflscape}  % Include the pdflscape package for landscape pages

\setlist{nolistsep}

\definecolor{mygray}{rgb}{0.5,0.5,0.5}
\lstdefinestyle{sparql}{
  backgroundcolor=\color{white},
  basicstyle=\ttfamily\footnotesize,  % Set the font size to small
  breakatwhitespace=false,
  breaklines=true,
  captionpos=b,
  commentstyle=\color{mygray},
  frame=single,
  keywordstyle=\color{blue},
  language=SQL,
  numbers=left,
  numbersep=5pt,
  numberstyle=\tiny\color{mygray},
  showspaces=false,
  showstringspaces=false,
  showtabs=false,
  stringstyle=\color{orange},
  tabsize=2,
  title=\lstname
}

\begin{document}

\title{Situated Ground Truths: Enhancing Bias-Aware AI by Situating Data Labels with SituAnnotate}
\author{Delfina S. Martinez Pandiani}
\authornotemark[1]
\email{dsmp@cwi.nl}
\orcid{0000-0003-2392-6300}
\affiliation{
  \institution{Centrum Wiskunde \& Informatica}
  \city{Amsterdam}
  \country{Netherlands}
  \postcode{1098 XG}
}

\author{Valentina Presutti}
\affiliation{
  \institution{University of Bologna}
  \city{Bologna}
  \country{Italy}
  \postcode{40127}
}

\renewcommand{\shortauthors}{Martinez Pandiani et al.}

\begin{abstract}

In the contemporary world of AI and data-driven applications, supervised machines often derive their understanding, which they mimic and reproduce, through annotations—typically conveyed in the form of words or labels. However, such annotations are often divorced from or lack contextual information, and as such hold the potential to inadvertently introduce biases when subsequently used for training. This paper introduces SituAnnotate, a novel ontology explicitly crafted for "situated grounding," aiming to anchor the ground truth data employed in training AI systems within the contextual and culturally-bound situations from which those ground truths emerge. SituAnnotate offers an ontology-based approach to structured and context-aware data annotation, addressing potential bias issues associated with isolated annotations. Its representational power encompasses situational context, including annotator details, timing, location, remuneration schemes, annotation roles, and more, ensuring semantic richness. Aligned with the foundational Dolce Ultralight ontology, it provides a robust and consistent framework for knowledge representation. As a method to create, query, and compare label-based datasets, SituAnnotate empowers downstream AI systems to undergo training with explicit consideration of context and cultural bias, laying the groundwork for enhanced system interpretability and adaptability, and enabling AI models to align with a multitude of cultural contexts and viewpoints.
\end{abstract}

\begin{CCSXML}
<ccs2012>

    <concept>
       <concept_id>10002951.10003152</concept_id>
       <concept_desc>Information systems~Information storage systems</concept_desc>
       <concept_significance>500</concept_significance>
       </concept>
   <concept>
       <concept_id>10002951.10003227.10003351</concept_id>
       <concept_desc>Information systems~Data mining</concept_desc>
       <concept_significance>500</concept_significance>
       </concept>
   <concept>
       <concept_id>10010147.10010178.10010187.10010188</concept_id>
       <concept_desc>Computing methodologies~Semantic networks</concept_desc>
       <concept_significance>500</concept_significance>
       </concept>
   <concept>
       <concept_id>10010147.10010178.10010187.10010195</concept_id>
       <concept_desc>Computing methodologies~Ontology engineering</concept_desc>
       <concept_significance>500</concept_significance>
       </concept>
    <concept>
       <concept_id>10010147.10010178.10010224</concept_id>
       <concept_desc>Computing methodologies~Computer vision</concept_desc>
       <concept_significance>500</concept_significance>
       </concept>
 </ccs2012>
\end{CCSXML}

\ccsdesc[500]{Information systems~Information storage systems}
\ccsdesc[500]{Information systems~Data mining}
\ccsdesc[500]{Computing methodologies~Semantic networks}
\ccsdesc[500]{Computing methodologies~Ontology engineering}
\ccsdesc[500]{Computing methodologies~Computer vision}

\keywords{situated AI, cultural bias, context, explainability, knowledge engineering}

% \received{15 September 2023}
% \received[revised]{31 Mar 2024}
% \received[accepted]{15 June 2024}

\maketitle

\section{Introduction}

In J. L. Borges' famous short story \textit{The Analytical Language of John Wilkins} \cite{borges1937analytical}, animals are classified into unconventional and seemingly bizarre categories.\footnote{Borges' story presents a fictitious taxonomy of animals, supposedly taken from an ancient encyclopaedia, which divies divides all animals into `` (a) those that belong to the emperor, (b) embalmed ones, (c) those that are trained, (d) suckling pigs, (e) mermaids, (f) fabulous ones, (g) stray dogs, (h) those that are included in this classification, (i) those that tremble as if they were mad, (j) innumerable ones, (k) those drawn with a very fine camel hair's brush, (l) others, (m) those that have just broken a flower vase, (n) those that resemble flies from a distance.'' } The story showcases the arbitrary and culturally-specific nature of categorization, a philosophical questioning into the complexities and subjectivity inherent in the act of classification. This theme finds a modern parallel in the rapid growth of artificial intelligence (AI) and data-driven applications, where classifying data is essential for training our machines \cite{qiu2016survey, beam2018big, salkuti2020survey}.

Labeled data, which underpins modern AI systems, is the result of vast processes of data annotation, where meaning is assigned most commonly through linguistic labels to data points. Given that data produced and annotated by humans possesses unique value, with the underlying belief that the ``human touch'' is indispensable to ensure accuracy and quality, the annotation process often depends on microlabor of human platform workers \cite{tubaro2020trainer}. 
Data annotation is deceptively complex, revealing a paradox where seemingly objective AI systems grapple with subjective annotations, resulting in inherent bias. This stems from the context-dependent nature of annotation, which challenges the notion of universal objectivity. In the digital age, AI systems, portrayed as objective, are constructed using data steeped in the subjectivity they aim to overcome. This intricate interplay between classification, subjectivity, AI data labeling, and bias emphasizes the complexities of modern AI development.

Data labeling processes are frequently shaped by human judgments, cultural viewpoints, and personal biases. It's important to clarify that the biases discussed in this paper should not be conflated with the "bias" term in machine learning models, which, mathematically speaking is an intercept or offset from an origin. Rather, we are focusing on cultural bias in the sense defined by \citet{APA2015cul}:

\begin{quote}
 \textit{the tendency to interpret and judge phenomena in terms of the distinctive values, beliefs, and other characteristics of the society or community to which one belongs. 
 % This sometimes leads people to form opinions and make decisions about others in advance of any actual experience with them (prejudice).
 }
\end{quote} 

This article delves into the technical aspects of accounting for cultural bias in the process of assigning semantic labels to data, with a case study of how this bias permeates the moment of labeling pixel areas of images within training datasets. This particular phase of human-led or human-evaluated annotation is critical, as the labels generated here become integral parts of input data for widely utilized models across various domains. Consequently, the ``data itself'' can harbor human biases, including stereotypes, prejudice, and racism. In this sense, this paper primarily concerns itself with the intersection of cultural and measurement bias, with measurement bias denoting faulty, low-quality, or unreliable data collection measures,  which can have many causes such as insufficient label options (e.g. binary gender \citep{scheuerman2019computers}) or from subjective views from labelers. These biases can lead to skewed interpretations and annotations, subsequently affecting the decisions made by AI systems. A compelling example of this phenomenon can be observed in the realm of computer vision, where popular datasets like ImageNet \citep{deng2009imagenet} play a pivotal role by providing ground truths or ``factual'' meanings to extensive image collections. Paradoxically, these datasets inadvertently reinforce contested political categories and cultural prejudices. For instance, an image of an indigenous person in traditional attire might be labeled as "half-naked," perpetuating a culturally biased perception as objective truth.  Classification frameworks hold authority in determining the significance of features, potentially amplifying specific worldviews while marginalizing others. Consequently, the ramifications extend beyond mere representation, encompassing the ability to mold societal perspectives and fortify preexisting biases.

The sway of these data biases is not limited to equity or fairness; it can significantly shape the very performance of AI systems reliant on them (e.g., \cite{davidson2019racial, wich2020impact}). Over the past decade, the issue of data bias has taken center stage \cite{ntoutsi2020bias, daneshjou2021lack}, with endeavors to “unbias” models and/or the data that they learn from have becoming a cornerstone in the pursuit of equitable AI systems \cite{chen2023bias}. However, any effort to encapsulate the intricate realities of the world inherently carries with it biases and perspectives rooted in context. In this sense, rather than the pursuit of defining and cultivating "unbiased" datasets—an increasingly improbable feat—a paradigm shift is emerging, which uses biased datasets with the awareness of this phenomenon, and tries to identify how bias affects results, embracing the nuanced, situated nature of annotations \cite{al2020identifying, ross2020measuring, wich2020investigating}. It is in this context that we propose the SituAnnotate ontology, a knowledge representation and capture tool poised to navigate the landscape of annotation situations for labeling data used to train AI systems.

\subsection{SituAnnotate to Enhance Cultural Bias-Aware AI} 

While considerable effort has been invested in establishing standards for capturing metadata pertaining to data and model production and reuse (e.g., data sheets \cite{gebru2021datasheets} and model cards \cite{mitchell2019model}, there is a lack of technical tools that allow both humans and machines to reason over such contextual information, a gap that persists especially at the level of singular annotations. This paper advocates for the explicit encoding of situational metadata alongside annotated data, so as to allow reasoning. This encoding should be designed to be both machine-readable and comprehensible/retrievable by human users.

This paper introduces SituAnnotate, an ontology-based module designed to formally represent the culturally-bound processes involved in annotating data. It builds upon the Description and Situations ontology design pattern \citep{gangemi2003understanding} to account for two key aspects: 1) the explicit tracking of culturally coded annotation situations, detailing how meaning is associated with data, and 2) the ability to reason over and compare annotations and their contexts. SituAnnotate,offers a structured and context-aware approach to annotating situational context, encompassing annotator type, temporal and spatial information, remuneration schemes, annotation roles, and more. SituAnnotate's core objective is to capture the contextual backdrop surrounding annotations while providing machine-readable representations of the circumstances in which data gains significance through linguistic labels. It builds upon the Dulce Ultralight ontology, ensuring robustness and consistency in knowledge representation, thereby facilitating the selection of specific data subsets based on annotation context criteria.

\subsection{Structure of the Paper} This paper is structured as follows: In Section \ref{related}, a review of related works is presented, covering AI data labeling practices, biases, and existing approaches to mitigate them. Section \ref{situ} introduces the SituAnnotate ontology, first describing the user requirement scenarios that guided the design of the ontology, and then defining fundamental concepts and design principles, and describing the core Classes. Section \ref{cv} discusses the case study of image annotations within computer vision pipelines. The evaluation protocol, including competency questions and results, are discussed in Section \ref{eval} The implications, contributions, and an example of module specialization of SituAnnotate are discussed in Section \ref{discussion}. Ultimately, Section \ref{conclusion} provides a concluding segment summarizing the key findings and the impact of SituAnnotate. The ontology is available online \footnote{\url{https://w3id.org/situannotate}} and documented in its GitHub repository.\footnote{\url{https://github.com/delfimpandiani/situAnnotate/}} The latter also contains the SPARQL queries and tests used for the evaluation of the ontology.

\section{Related Work}
\label{related}

\subsection{Annotated Data Hunger}
The significance of data in the realm of machine learning cannot be overstated. As \cite{safdar2020ethical} succinctly puts it, "ML is data-hungry. Deep learning is data-ravenous." To effectively train supervised models, datasets with meticulously annotated labels are imperative, as they furnish the necessary supervised information to guide model training and estimate functions or conditional distributions over target variables from input data. Nevertheless, the process of manually labeling data can be labor-intensive and time-consuming. In response to this challenge, there are alternatives such as pseudo-labelling and label propagation, as discussed by \cite{wang2023scientific}, which offer the possibility of automatically annotating extensive unlabelled datasets based on a limited set of accurate annotations. This process then makes available ground truths an indispensable foundation for reliable model performance assessment and validation.

\subsection{The Human Touch in Annotated Data}

Data annotation, as highlighted by \cite{tubaro2020trainer}, predominantly relies on human involvement, recognizing the unique value attributed to data produced and annotated by humans. This underscores the crucial role of the "human touch" in ensuring the accuracy and quality of annotated data. Geiger et al.'s work \cite{geiger2020garbage, geiger2021garbage} offers a comprehensive review of the landscape of human labeling of training data in machine learning, delving into best practices in this field. They argue that much of this labeling work aligns with structured content analysis, a methodology supposed to be ``systematic and replicable’’ \cite[p. 19]{riffe2019analyzing} and historically employed in the humanities and social sciences to transform qualitative or unstructured data into categorical or quantitative data. This structured content analysis entails the work of ``coders’’ or ``labelers’’ who individually assign labels or annotations to items in the dataset according to ``coding schemes’’, after which inter-rater reliability is assessed. Historically undertaken by students, crowdwork platforms like Amazon Mechanical Turk have become most common for data labeling tasks, with new platforms emerging to support micro-level labeling and annotation, including, for example, citizen science initiatives where volunteers collaborate to label data across various domains (e.g., \cite{chang2017revolt}).

\subsection{The Garbage In, Garbage Out Principle}

In the realm of machine learning, the axiom ``garbage in, garbage out'' \cite{babbage1864passages} reverberates as a familiar cliche, emphasizing that the quality of data used in a process directly influences the quality of the outcomes. Garbage data extends to include not only inaccuracies but also decontextualized or biased information that lacks relevant connections or meaning. Data quality concerns are often overlooked in ML research and education \cite{geiger2021garbage}, but it is essential for those applying ML in real-world domains to grasp the implications of low-quality or biased training data. The idea that automated systems are not inherently neutral and instead reflect the priorities, preferences, and prejudices of those who have the power to mold artificial intelligence is an increasingly public topic of discussion, especially given that many datasets have been found to be systematically biased along various axes, including race and gender, which impacts the accuracy of those ML model. For example, \citep{buolamwini2018gender} investigates the false assumption of machine neutrality, and the \textit{coded gaze}--the algorithmic `way of seeing' which classifies content through researcher- and machine-labeled categories--which ``reflects both our aspirations and our limitations'' \cite[p. 44]{buolamwini2017gender}. Another example is how the geographical sampling of Flickr images as well as the use of English as the primary language for dataset construction and taxonomy definition result in inherent cultural bias within the datasets \citep{de2019does}, with work being done to design new annotation procedures that enable fairness analysis \citep{schumann2021step}. As such, evaluating supervised models solely with a held-out subset of the training data can obscure systematic flaws, especially in cases where the model is used for contentious decisions like those in finance, hiring, welfare, and criminal justice. 
% Subtle issues arise, exemplified by a study attempting to classify criminals from noncriminals using facial images, where problematic labels were derived from mug shots and professional social network profiles, essentially creating a "smile classifier" instead of a criminality classifier, as critiqued by \cite{bergstrom2021calling}.

\subsection{Identifying and Documenting Bias in Data}

AI research often relies on biased perspectives in ground truth datasets, potentially causing issues when lacking proper context. New frameworks aim to clarify the assumed knowledge within datasets and deployed AI systems to combat this problem.

\subsubsection{De-biasing ML}
There are efforts to ``de-bias’’ ML (surveys by \cite{mehrabi2021survey, friedler2019comparative}), including via developing domain-independent fairness metrics to test and modify trained models or predictions. For example, \cite{ghai2022d} addresses the issue of social biases in AI algorithms by proposing D-BIAS. D-BIAS is a visual interactive tool that employs a human-in-the-loop AI approach to audit and mitigate social biases from tabular datasets. D-BIAS uses graphical causal models to represent relationships among features in the dataset and inject domain knowledge. Users can detect bias against specific groups, such as females or black females, and refine causal models to mitigate bias while minimizing data distortion. Other approaches have been through dataset preprocessing \cite{calmon2017optimized} or database repair \cite{salimi2020database}.

\subsubsection{Documenting (Meta)Data}
Other efforts have designed standards for capturing metadata pertaining to data and model production and reuse. \cite{mitchell2019model} propose the use of ``model cards'' to accompany trained machine learning models, which are concise documents that provide benchmarked evaluations of models under various conditions. These cards also disclose the intended use cases, evaluation procedures, and relevant information about the model. \cite{gebru2021datasheets} introduce the concept of ``datasheets for datasets'' drawing an analogy to datasheets for electronic components. They propose that every dataset should be accompanied by a datasheet that documents its motivation, composition, collection process, recommended uses, and more. This approach facilitates better communication between dataset creators and consumers, prioritizing transparency in data collection. Other approaches include “data statements” \cite{bender2018data}, “nutrition labels” \cite{holland2020dataset}, a “bill of materials” \cite{barclay2019towards}, “data labels” \cite{beretta2018ethical} and “supplier declarations of conformity” \cite{arnold2019factsheets}. Additionally, \cite{jo2020lessons} argue for the importance of a new specialization within machine learning focused on methodologies for data collection and annotation. They draw parallels with archival practices, where scholars have developed frameworks and procedures to address challenges like consent, power, inclusivity, transparency, ethics, and privacy. By incorporating these approaches from archival sciences, they encourage the machine learning community to be more systematic and cognizant of data collection, particularly in sociocultural contexts.

\subsubsection{Investigating  Annotator Bias}
Moreover, efforts to enhance transparency and accountability in the machine learning community have focused on detecting and addressing annotator bias. \cite{wich2020impact} identify annotation bias by analyzing similarities in annotator behavior. To achieve this, they construct a graph based on annotations from different annotators, apply a community detection algorithm to group annotators, and train classifiers for each group to compare their performances. This approach enables the identification of annotator bias within a dataset, ultimately contributing to the development of fairer and more reliable hate speech classification models. Within the context of hate speech detection systems, \cite{al2020identifying} delve into the issue of annotator bias with a specific focus on demographic characteristics. They construct a graph based on annotations from various annotators and utilize community detection algorithms to group annotators based on demographics. They then proceed to train classifiers for each demographic group and conduct performance comparisons. This rigorous approach enables them to shed light on how demographic features like first language, age, and education significantly correlate with performance disparities. 

\subsection{Evolving Perspectives on Relationality in Data Annotation}

Recent attention has shifted towards viewing data annotation processes as intricate amalgamations of technical considerations and sociocultural insights. This interdisciplinary approach finds fertile ground in research streams, notably within Natural Language Processing (NLP). Key works such as those by \citep{basile2021toward, diaz2022accounting, hovy2021importance} delve into the nuanced implications of annotating data within specific sociocultural contexts, illuminating the subjective elements inherent in annotation practices.

Hovy et al. \cite{hovy2021importance} argue convincingly that the limitations of current NLP systems arise from a myopic focus on information content, neglecting the crucial social dimensions of language. Their research demonstrates how NLP systems falter in interpreting social aspects of language, such as sarcasm and irony, which demand a nuanced understanding beyond mere message content. Proposing a transformative shift akin to the evolution seen in behavioral economics, they advocate for integrating human and social factors into NLP, opening avenues for more robust and equitable tools.

In a similar vein, Basile et al. \cite{basile2021toward} propose a groundbreaking paradigm called \textit{data perspectivism}, challenging conventional approaches to data annotation in machine learning. Their advocacy for incorporating diverse opinions and perspectives of human subjects marks a departure from traditional annotation methods, particularly beneficial in subjective tasks like those within NLP. Furthermore, Diaz et al. \cite{diaz2022accounting} emphasize the need for a relational understanding of offensive language labeling and detection, underscoring its subjective nature and sociocultural contexts. By exploring examples from marginalized communities, they illuminate how offensive speech can function as a form of resistance against oppressive social norms. Their insights highlight the importance of considering broader social contexts in data annotation, urging annotators to provide contextual cues for improved quality and reliability.

\subsection{Ontologies for Digital Hermeneutics}

Ontologies formally represent data semantics in a machine-readable format, enabling explicit semantics and facilitating queries based on concepts and relationships \citep{bannour2011towards}. Previous research has applied ontology-driven approaches in fields like image understanding and computer vision, especially in addressing the challenge of image interpretation. Notably, the IECA ontology provides a model for representing interpretative encounters between human observers and cultural artifacts, exploring content, context, and relationships among alternative interpretations of cultural objects \citep{ieca-ontology}. A recent work focuses on modeling interpretation and meaning for art pieces, presenting a data model for describing iconology and iconography. Additionally, the Historical Context Ontology (HiCO) aims to outline relevant issues related to the workflow for stating and formalizing authoritative assertions about context information for cultural heritage artifacts \citep{daquino2015historical}. Also, the VIR (Visual Representation) ontology, constructed as an extension of CIDOC-CRM, sustains the recording of statements about the different structural units and relationships of a visual representation, differentiating between object and interpretative act \citep{carboni2019ontological}. These developments illustrate the versatility of ontologies in addressing various interpretation challenges in different domains.

\section{SituAnnotate}
\label{situ}

\subsection{Situating (Ground) Truths}

This paper contends that a crucial step towards the goal of responsible and ethical AI \cite{crawford2021atlas} involves the deliberate grounding of assumed objective truths within their respective situated contexts. This view aligns with the growing need for technical solutions to challenge the conventional notion of an unequivocal truth in human annotation \cite{aroyo2015truth}, to adopt a power-aware approach to data design and production \cite{miceli2022studying}, and to reveal how AI, ML, and data practices inadvertently perpetuate colonial power dynamics and value systems \cite{birhane2020algorithmic, mohamed2020decolonial}.

We philosophically adhere to the idea of ``situated grounding'' in training data, echoing the concept of ``situatedness'' exemplified by Donna Haraway in 1988. Haraway challenges the traditional detached view of vision, characterized as a "conquering gaze from nowhere":

\begin{quote}
\textit{This is the gaze that mythically inscribes all the marked bodies, that makes the unmarked category claim the power to see and not be seen, to represent while escaping representation}. \cite[p. 581]{haraway1988situated}
\end{quote}

We are inspired by Haraway's alternative paradigm of ``situated knowledges,'' advocating for a perspective rooted in complex, contradictory, structured bodily experiences, rather than from an assumed objective standpoint \cite[p. 589]{haraway1988situated}. Embracing this paradigm involves recognizing the multifaceted nature of localized knowledge.

\subsection{Ontology Requirement Elicitation via Scenarios}

\subsubsection{Scenario-Based Requirement Identification}

Due to the significant growth of ontologies attributable to the Semantic Web, numerous methodologies and tools have emerged to support their development. The ontology development cycle shares both similarities with the software development lifecycle, including the specification of requirements \cite{bezerra2013evaluating}. We follow the eXtreme Design (XD)  ontology engineering methodology \citep{presutti2009extreme}, which  incorporates project initiation steps, requirements analysis, development, testing and release of an ontology. Specifically, we refer to its use of scenarios, which serve as a valuable tool for eliciting requirements in ontology development. Scenerios are natural language descriptions that outline ordered processes and activities, providing a structured pathway for modeling knowledge \cite{suarez2015neon}. Scenarios encapsulate typical situations, facilitating a systematic approach to ontology construction. By leveraging a scenario-based approach to identify requirements and competency questions (CQs) for ontology evaluation (see Section \ref{eval}), we can effectively contextualize these elements in terms of their relevance, origin, and applicability. This approach grounds the ontology by establishing a clear link between scenarios, CQs, and real-world challenges or information needs. Through this integration, the ontology evaluation process becomes more robust and well-founded, ensuring that it adequately addresses the underlying objectives and requirements of the intended application domain. In the context of developing the SituAnnotate ontology, scenarios serve as blueprints for addressing needs regarding the contextualization of annotations, aligning the ontology with research objectives. These scenarios were derived from conversations with academic researchers from multiple disciplines related to data annotation, including computer vision, natural language processing, knowledge engineering, and ethical and responsible AI. The focus was on identifying the types of knowledge deemed valuable for labeling research objects within these scholarly domains. From these dialogues, a total of 11 scenarios emerged, reflecting a diverse range of research-based annotation processes. Although primarily catering to research-oriented endeavors, these scenarios also hold relevance for product-related annotations, albeit to a lesser extent.

\subsubsection*{Identified Scenarios}

The SituAnnotate ontology is inspired by Donna Haraway's `situated knowledges' paradigm, emphasizing context-dependent perspectives over detached objectivity. To ensure its effectiveness, the following 11 user requirement scenarios were identified, serving as practical examples of the intricate challenges SituAnnotate addresses. These scenarios highlight that ground truths are context-dependent, nuanced entities:

\vspace{1em}
% _____Scenario 1: Geographic Distribution of Annotation Situations__________________
\noindent \textbf{Scenario 1: Geographic Distribution of Annotation Situations}

\noindent \textit{I want to understand the geographic distribution of annotation situations in SituAnnotate. Specifically, I want to know which countries have been the location of annotation situations, how many annotation situations were located in each country, and which country has the highest number of annotation situations.}

\noindent \textbf{Rationale}: This scenario aims to shed light on the geographic scope of annotation situations captured by SituAnnotate. Understanding where annotation activities are concentrated can provide insights into regional preferences, data availability, and potential biases in the annotation process.

\vspace{1em}
% _______________________
\noindent \textbf{Scenario 2: Temporal Filtering of Annotation Situations}

\noindent \textit{I want to research the temporal aspects of annotation situations. Specifically, I want to select a specific period of time and identify which annotation situations a particular image has been involved in during that time. This allows me to track the history of annotations for the image and observe how they may evolve over time.}

\noindent \textbf{Rationale}: This scenario tests SituAnnotate's ability to track temporal information, enabling precise filtering based on annotation dates. This feature also facilitates the comparison of annotations before and after significant cultural moments, such as the COVID-19 pandemic, offering insights into how labels for the same image may evolve over time in response to societal changes.

\vspace{1em}
% _______________________
\noindent \textbf{Scenario 3: Remuneration Schemes in Annotation Situations}

\noindent \textit{For a certain dataset, I want to know which remuneration schemes have been used in annotation situations meant to create annotations for it.}

\noindent \textbf{Rationale}: This scenario explores the various compensation models employed in annotation situations that have led to annotations for a specific dataset. Identifying remuneration schemes informs us about the motivations and incentives driving annotators, which can impact the quality and consistency of annotations.

\vspace{1em}
% _______________________
\noindent \textbf{Scenario 4: Annotated Entity Types in Annotation Situations}

\noindent \textit{I want to gain insights into the types of entities that have been annotated within the SituAnnotate ontology. Specifically, I want to know the categories of entities, such as images or documents, that have undergone annotation and are represented in the SituAnnotate knowledge graph.}

\noindent \textbf{Rationale}: This query illuminates the entities whose annotations have been integrated into the SituAnnotate ontology. It offers insight into the categories of entities, such as images and documents, that have undergone annotation and are represented within the SituAnnotate knowledge graph. This comprehension is crucial for domain-specific applications as it unveils the breadth of concepts encompassed by the ontology.

\vspace{1em}
% _____Scenario 5: Identifying Annotations based on Lexical Entry___________________
\noindent \textbf{Scenario 5: Identifying Annotations based on Lexical Entry}

\noindent \textit{I want to identify all entities that have been annotated using a specific lexical entry, such as "surfboard." Additionally, I want to know the roles that these annotations serve.}

\noindent \textbf{Rationale}: This question exemplifies how the ontology can be leveraged for the identification of all entities, or entities of a specific type (e.g., images), that have been annotated with the same lexical entry (e.g., "surfboard") and the corresponding annotation roles (e.g., detected object). This query is instrumental in gaining insights into the usage and impact of specific lexical entries across various annotations.

\vspace{1em}
% _______________________
\noindent \textbf{Scenario 6: Identifying Contextual Information for Annotations}

\noindent \textit{For a specific situation in which a lexical entry was used to annotate an entity, I want to know the contextual factors associated with the annotation situation, including the country, date, annotated dataset, remuneration scheme, detection threshold, and details about the annotator.}

\noindent \textbf{Rationale}: This question aims to provide comprehensive context for a particular annotation scenario, encompassing geographical and temporal aspects, the dataset under annotation, remuneration specifics, detection thresholds, and annotator attributes. It offers a powerful tool for understanding how a ground truth is situated within its originating context.

\vspace{1em}
% _______________________
\noindent \textbf{Scenario 7: Filtering Annotations by Reliability and Roles}

\noindent \textit{I want to filter annotations based on their reliability and roles. Specifically, I want to identify entities with annotations classified under specific annotation roles, such as detected objects or detected emotions, with annotation strengths exceeding certain thresholds. Additionally, I want to know the labels assigned to these entities.}

\noindent \textbf{Rationale}: This question delves into annotations categorized by specific roles (e.g., detect object, detected emotion, detected action) and their associated annotation strengths. It allows for the filtering of entities based on the reliability or strength of annotations and provides insight into the specific labels assigned to these entities.

\vspace{1em}
% _______________________
\noindent \textbf{Scenario 8: Identifying Concepts Typing Annotations about Entities}

\noindent \textit{I want to identify the underlying concepts that give meaning to the words used in annotating a specific entity. Specifically, I want to understand the conceptual word senses linked to that entity by categorizing the assigned labels. Additionally, I aim to determine these concepts' roles and the strength of their association with the annotated entity.}

\noindent \textbf{Rationale}: This question delves into the conceptual basis of annotations for a specific entity by focusing on the conceptual word senses sourced from linguistic resources like ConceptNet, WordNet or FrameNet. These resources categorize the lexical units used in annotations, such as when the word ``impressionism'' is employed as a label for an image, it can be typed by the concept \url{https://www.conceptnet.io/c/en/impressionism}. The aim is to uncover associated concepts and their roles (e.g., detected emotion or detected action), along with the strength of these connections. By providing insights into the semantic underpinnings of annotations, this approach enhances our understanding of their reliability and significance.

\vspace{1em}
% _____Scenario 9: Tracking Annotators Responsible for Annotation Labels_________________
\noindent \textbf{Scenario 9: Tracking Annotators Responsible for Annotation Labels}

\noindent \textit{I want to identify the annotators responsible for specific labels associated with a particular image. Specifically, I want to attribute annotations to individual annotators, enabling an assessment of their contributions to the annotation process.}

\noindent \textbf{Rationale}: This scenario delves into the identification of the annotators accountable for specific labels associated with a particular image. This level of detail enables the attribution of annotations to individual annotators, facilitating an assessment of their contributions.

\vspace{1em}
% _______________________
\noindent \textbf{Scenario 10: Artificial Annotators and Shared Model Architectures}

\noindent \textit{I want to explore artificial annotators with shared model architectures within SituAnnotate. Specifically, I want to know what types of annotations about an entity were created by artificial annotators with a specific model architecture. Additionally, for each of these annotators, I want to determine the dataset they were pretrained on, if applicable.}

\noindent \textbf{Rationale}: This question explores artificial annotators that employ a shared architectural backbone for creating annotations of various types. Identifying shared model architectures sheds light on the integration of automated annotation tools within annotation pipelines. Additionally, it provides insights into the prevalence of specific model architectures and their pretraining on various datasets, contributing to a broader understanding of automated annotation methods.

\vspace{1em}
% _____Scenario 11: Identifying Image Caption Annotations and Annotators__________________
\noindent \textbf{Scenario 11: Identifying Image Caption Annotations and Annotators}

\noindent \textit{I want to focus on image caption annotations and the annotators responsible for them. Specifically, I want to identify the caption annotations for a specific image and determine who the annotators are for each caption annotation.}

\noindent \textbf{Rationale}: This query focuses on revealing caption annotations and their respective annotators for a given image. It is vital for examining the generation and attribution of textual descriptions, shedding light on the creators of these annotations and their role in conveying information about the image.

\subsection{SituAnnotate's Core Concepts}

\begin{figure*}[h]
  \includegraphics[width=\textwidth]{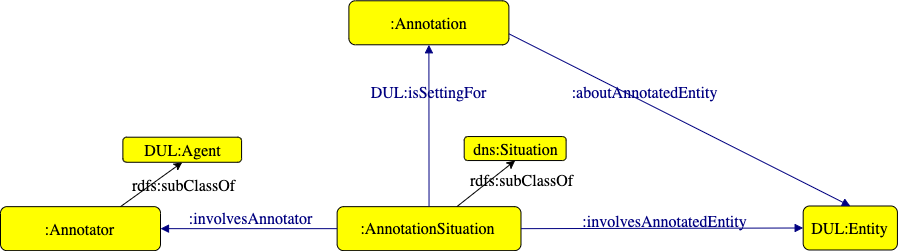}
  \caption{SituAnnotate at a glance: Core concepts connecting annotations, annotation situations, and annotators.}
  \label{fig:overview}
\end{figure*}

Based on the requirements elicited, the core goal of the SituAnnotate ontology is to \textit{situate} annotations by connecting them not only to the entity that they describe, but also to the general situation and to the annotator and other factors involved in it. The SituAnnotate ontology was designed to provide a structured and context-aware representation of annotation situations and their associated entities. As such, the three core Classes of the ontology are \textit{Annotation}, \textit{AnnotationSituation}, and \textit{Annotator} (see Figure \ref{fig:overview}). 

\subsection{Aligning with the Dolce Ultra Light (DUL) Ontology}

To ensure the robustness and consistency of the SituAnnotate ontology, it draws inspiration from and aligns with the Dolce Ultra Light (DUL) ontology. DUL is a foundational ontology inspired by cognitive and linguistic considerations, which aims to model a commonsense view of reality and provide general categories and relations needed for coherent ontological modeling \citep{borgo2022dolce}. DUL takes into account the requirements from semantic web modeling practices, the need for simplified semantics as in natural language processing lexicons, and the need for some extensions of DOLCE categories, by reusing the D\&S (Description and Situations) ontology
pattern framework \citep{gangemi2003understanding}. DnS supports a first-order manipulation of theories and models, and was chosen as a core design pattern of SituAnnotate because it allows for the modeling of non-physical objects, such as social concepts, whose intended meaning results from statements, i.e. they arise in combination with other entities. Specifically, DnS axioms capture the notion of situation as a unitarian entity out of a State of Affairs (SoA), that
is constituted by the entities and the relations among them, and a description as an entity that partly represents a (possibly formalized) theory that can be conceived by an agent. By aligning with DUL, the SituAnnotate ontology benefits from a well-established framework that enhances the ontological modeling of situations, entities, and their relationships. This alignment also promotes interoperability with other ontologies, enabling broader use and integration with existing semantic resources.

\subsection{Annotation Situations and Contextual State of Affairs}

\begin{figure*}[h]
  \includegraphics[width=\textwidth]{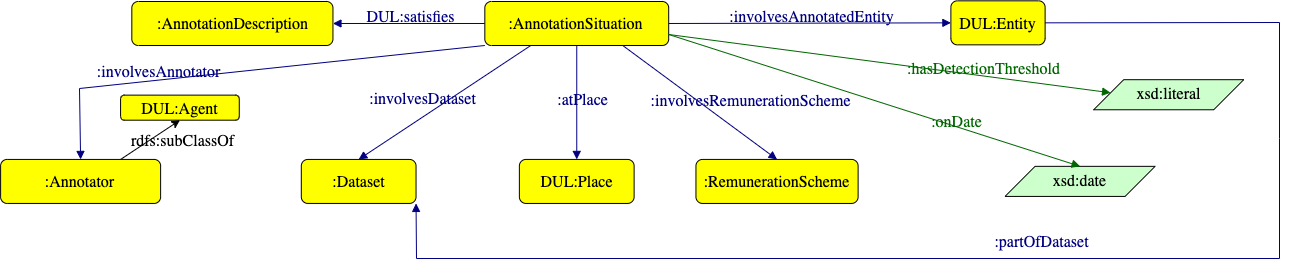}
  \caption{A detailed view of the SituAnnotate Ontology's core building block, the AnnotationSituation class.}
  \label{fig:as}
\end{figure*}

The first central Class of the SituAnnotate ontology is the \textit{AnnotationSituation} class, depicted in detail in Figure \ref{fig:as}. This class functions as the cornerstone of the ontology, a subclass of the DUL \textit{Situation} class to maintain alignment with DUL's situational modeling. In essence, the \textit{AnnotationSituation} encapsulates the comprehensive state of affairs in which an annotation may occur: at a precise moment in time, at a specific \textit{Place}, potentially involving a specific \textit{Annotator}, a \textit{RemunerationScheme} by which she is paid for her labor, a \textit{Dataset} to which the \textit{AnnotatedEntity} belongs, and more.  Furthermore, each \textit{AnnotationSituation} must adhere to an \textit{AnnotationDescription}. Additionally, this class can incorporate other pertinent details unique to the situation. By serving as a representation of the contextual environment in which annotations transpire, the \textit{AnnotationSituation} class interconnects all pertinent data, whether contextual or otherwise, associated with the annotation process.

\subsection{Annotations and Annotation Roles}

\begin{figure*}[h]
  \includegraphics[width=\textwidth]{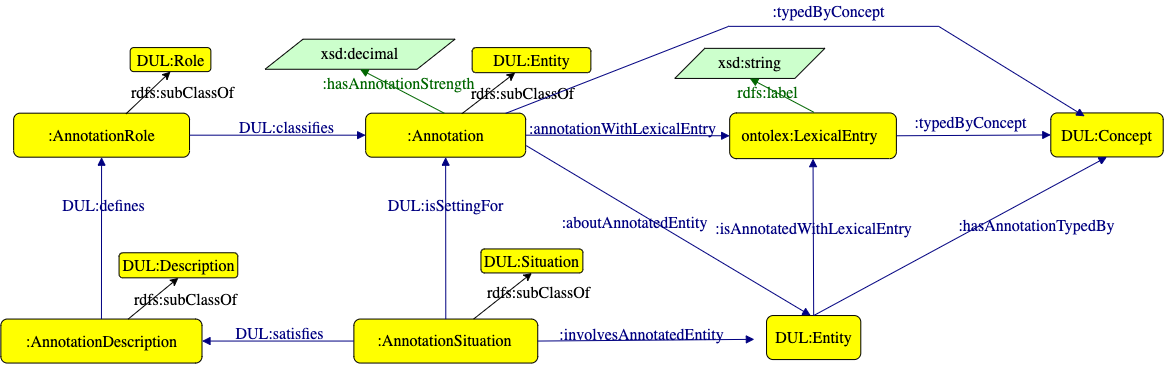}
  \caption{Deep Dive into the Annotation class: annotation instances connect annotated entities to lexical entries by fullfilling a specific AnnotationRole in a certain AnnotationSituation.}
  \label{fig:annotation}
\end{figure*}

The second core class is \textit{Annotation}. Instances of this class represent the units responsible for attaching specific meanings, conveyed by lexical units, to an annotated entity in the context of a particular \textit{AnnotationSituation} (see Figure \ref{fig:annotation}). Annotations are 
classified by their \textit{AnnotationRole}, a subclass of \textit{Role}. These roles are defined within \textit{AnnotationDescriptions}, adding semantic richness to the ontology, thus enhancing its expressiveness and precision. This approach allows for the representation of diverse annotation types and their roles within the annotation process. Notably, SituAnnotate introduces a distinctive feature where an \textit{Annotation} is a first-order instance capable of establishing relationships with other instances, extending beyond mere textual labels (e.g., "woman," "happiness," or "cemetery"). Instances of the \textit{Annotation} class are not only linked to their corresponding lexical entries but also to the \textit{AnnotatedEntity} they describe (e.g., an image), the specific annotation role they fulfill (e.g., "detected object," "detected emotion," or "detected scene"), the concept typing the lexical entry (e.g., conceptnet:woman), and, importantly, the \textit{AnnotationSituation} within which the annotation originated. This interconnection enables explicit queries to determine the context in which a specific entity was associated with a particular lexical label.

\subsection{Annotators}

The third key class in SituAnnotate is \textit{Annotator}. In this ontology, an Annotator can take one of two forms: an \textit{ArtificialAnnotator}, representing automated programs utilizing a specific \textit{ModelArchitecture} pretrained on a designated \textit{Dataset}, or a \textit{HumanAnnotator}. The \textit{HumanAnnotator} category is further subdivided into two subclasses: \textit{IndividualHumanAnnotator} and \textit{HumanAnnotatorCommunity}. This differentiation was introduced to handle situations where gathering specific demographic information about individual annotators might be challenging due to privacy considerations. In these cases, data is anonymized by combining and presenting averages or other statistical metrics.

When initially selecting attributes for annotators, we prioritized those most relevant to annotator demographics and their impact on labeled data. Attributes such as  \textit{PoliticalAffiliation}, \textit{ReligiousAffiliation}, \textit{IndigenousAffiliation}, and country of upbringing were chosen to reflect sociocultural contexts, particularly in the context of global crowdsourcing tasks. This selection was also influenced by the need to showcase the ontology's capability to capture diverse sources and methods used in assigning meaning to entities. While the ontology currently accommodates a small number of annotator attributes, it is designed to be flexible and scalable. Additional attributes, such as gender, sexual orientation, class, and migration background, can be seamlessly integrated into the ontology's extensible class hierarchies or modular design. Annotator communities, created by amalgamating data from annotator sets for privacy protection, can also be associated with affiliations using the ``predominant" version of affiliation relationships. In essence, this formalization allows for the comprehensive representation of various annotators employed to attribute meaning to an entity using a lexical label. This flexibility enhances the ontology's capacity to capture the diverse sources and methods used in assigning meaning to entities, including computer vision models, individual annotators, or annotator communities (e.g., the collective annotation provided by the Imagenet dataset annotators). 

However, the inclusion of new attributes should be approached with careful consideration, especially regarding ethical factors. Privacy concerns surrounding sensitive demographic information, such as gender, sexual orientation, and migration background, must be thoroughly addressed to protect the privacy and confidentiality of annotators. Additionally, the potential for biases and discrimination in the collection and use of demographic data should be carefully evaluated to ensure fairness and equity in annotation practices. Data sparsity and the need for validation and standardization to maintain data integrity are also critical ethical considerations.

\begin{figure*}[h]
  \includegraphics[width=\textwidth]{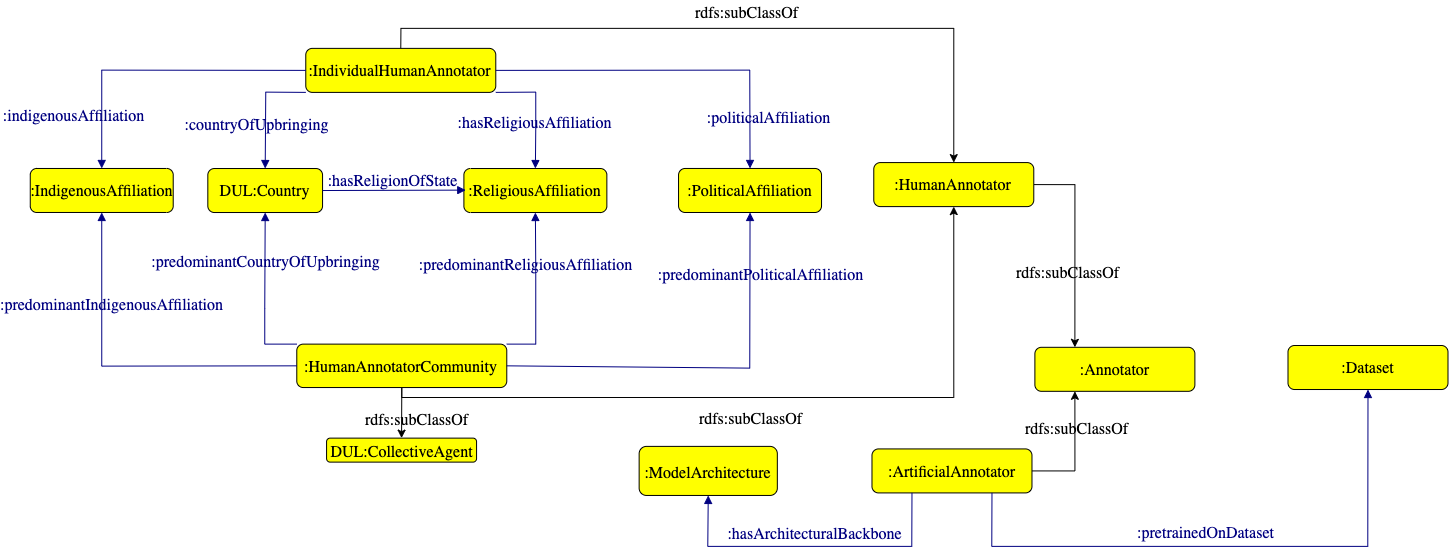}
  \caption{Deep Dive into the Annotator class: SituAnnotate allows the formal representation of different types of annotators and relevant characteristics that may influence their annotation choices.}
  \label{fig:annotator}
\end{figure*}

\section{Case Study: Image Annotation Situations}
\label{cv}
To harness the full potential of our ontology and later assess these scenarios, we expanded our work into a case study focusing on image annotations. These annotations are pivotal in computer vision, a field that stands to benefit significantly from SituAnnotate and our contributions. Computer vision heavily relies on assigning meaning through labels to images, making it particularly susceptible to biases, including human, algorithmic, and interpretational biases \citep{laranjeira2021}. Thus, computer vision serves as an ideal case study, highlighting its heavy reliance on labels and its pronounced vulnerability to concealed biases.

\subsection{Motivation}

In the realm of computer vision, image annotation labels are of paramount importance, serving as the linchpin for understanding, retrieving, and managing the burgeoning volumes of images \citep{llamas2016applying, saleh2018image, wang2021data}. These labels, often structured and endowed with semantic meaning through label- and graph-based resources, bridge the chasm between raw image content and its comprehension. Particularly in complex image scenes, the semantic annotation of objects within them empowers automatic understanding and interpretation \citep{sager2021survey}. Increasingly, linguistic resources and graphs like WordNet \citep{miller1995wordnet}, ConceptNet \citep{liu2004conceptnet} and Framester \citep{gangemi2016framester} are used to assign and organize labels that give meaning to the raw content of images, for example in the form of scene graphs (e.g., Visual Genome \citep{krishna2017visual}) or taxonomies (e.g., the Tate collection\footnote{\url{https://github.com/tategallery/collection/issues/27}}). These amplify the semantic richness of image features, bolstering image labeling and retrieval systems \citep{martinez2021automatic, samih2021semantic, tariq2017learning}. 

Critically, the structured representations arising from these annotations also double as invaluable ground truths for the training of computer vision systems, contributing substantially to their precision and efficacy. However, it's imperative to acknowledge that the meanings attributed to images do not exist in a cultural vacuum. Images communicate concepts through a fusion of raw features like lines, colors, shapes, and sizes, alongside culturally coded elements, an aspect that Roland Barthes termed 'connotation' \citep{barthes1980camer1a}. These coded elements guide human decision-making regarding object identification, labeling, feature ascription, and relationship establishment. In essence, the extraction and portrayal of semantic elements from visual content constitute a code system intricately intertwined with cultural context. This is because \textit{visuality}, different from the purely biological process of vision, is flexible and encompasses ``the way that we encounter, look at, and interpret images based on the social, cultural, technological, and economic conditions of their viewing'' \cite[32]{giotta2020waysseeing}. That is, visuality is a cultural practice with a history marked by different habits or ways of seeing, as well as different types of spectators \citep{foster1988preface}. This cultural context remains embedded within computer vision pipelines, persisting even in ostensibly straightforward processes like object detection. The 'distant viewing' framework, as introduced by \citet{arnold2019distantviewing}, emphasizes the indispensability of a culturally and socially constructed code system to render the semantics of visual content explicit. Labeling and classification systems, though seemingly objective, can inadvertently mirror the values of specific groups or cultures, thereby centralizing power within the process. Despite these intricacies, there lingers a prevailing faith in the objectivity of image labels found in benchmark datasets, often underestimating the cultural and subjective nature of image annotation \citep{giotta2020waysseeing}.

\subsection{The Image Annotation Situation Specialization}

\begin{figure*}[h]
  \includegraphics[width=\textwidth]{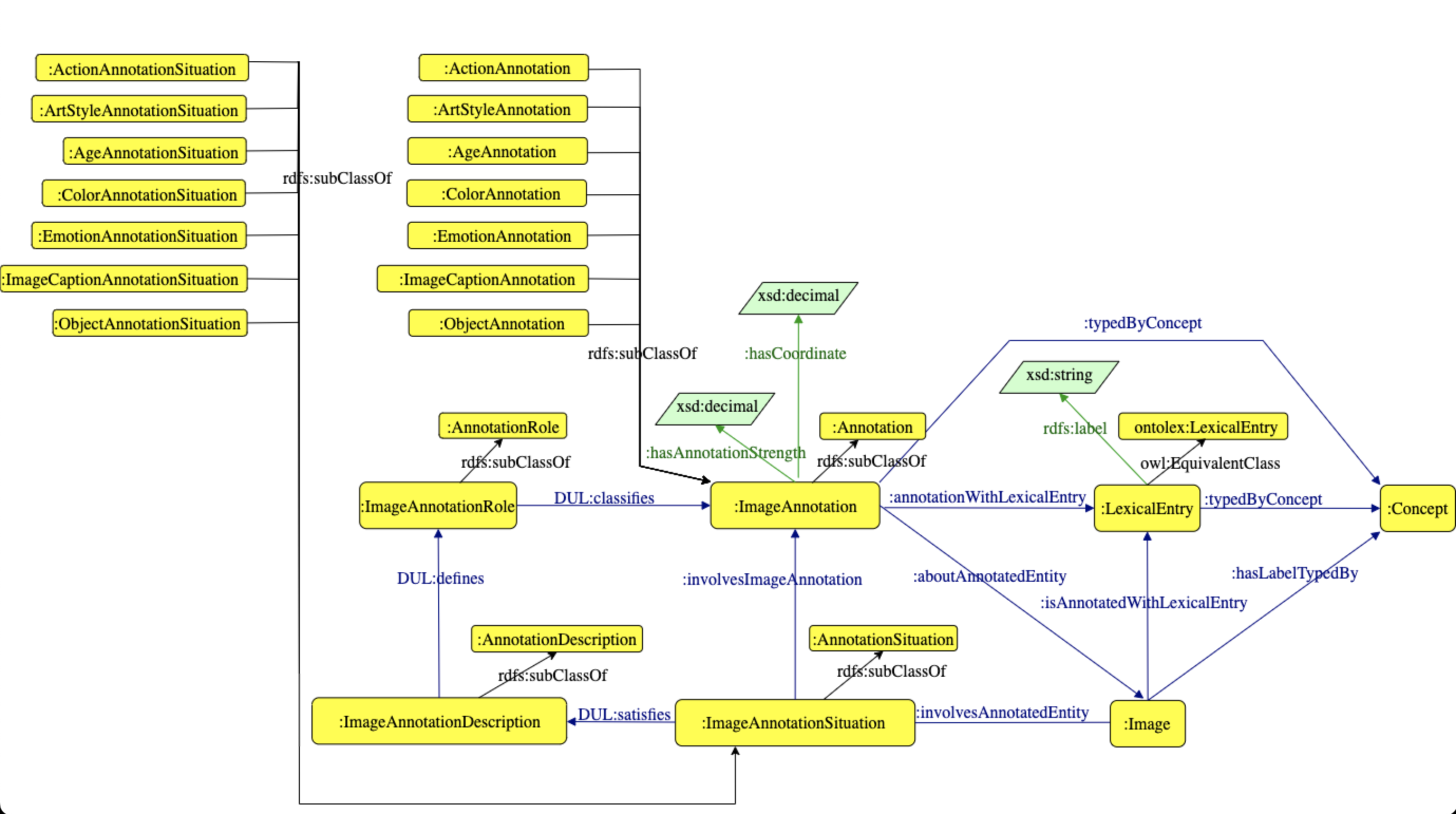}
  \caption{Specialization of the SituAnnotate pattern specifically for Image Annotation Situations (IAS), crucial in the field of Computer Vision (CV). Further modular specializations can be applied to capture details specific to certain types of annotation situations, such as object detection.}
  \label{fig:ias}
\end{figure*}

We've specialized SituAnnotate to create the Image Annotation Situations (IAS) module, depicted in Figure \ref{fig:ias}, with the explicit purpose of tracing the origins of image meanings within culturally coded annotation contexts and facilitating their comparison. This approach is rooted in the notion that an image's semantic labels depend on the specific annotation situation under which it is interpreted. In the IAS module, image annotation is recognized as a contextual situation, similarly to \cite{vacura2008describing}, represented by the class \texttt{ImageAnnotationSituation}. This context encapsulates all entities relevant to the annotation process, including the image, annotator, annotation time, location, remuneration details, dataset creation purpose, and more. By applying the Situation pattern, the \texttt{ImageAnnotationSituation} class provides a structured framework for contextualizing these entities, allowing for shared features such as location, time, view, causality, and systemic dependencies to be captured.

The IAS module emphasizes the need to describe an annotation situation through an \texttt{ImageAnnotationDescription}. This description defines the roles and concepts that participate in the state of affairs. The IAS module not only incorporates cultural contextual information regarding image annotation situations but also facilitates comparison between different annotation situations associated with the same image object. This enables users to query and analyze the contexts in which potentially contradictory interpretations of the same image were produced.

Furthermore, SituAnnotate's IAS module already includes classes to support various types of annotations and annotation situations, such as art style annotations, color annotations, object annotations, action annotations, emotion annotations, caption detection, and more. Furthermore, the ontology accommodates detailed annotations, including the assignment of labels to specific regions within an image using the property \texttt{:hasCoordinate}. This feature enables the representation of bounding box annotations for pixels within an image.

Additionally, SituAnnotate offers different image annotation descriptions for the mentioned annotation types (e.g., emotion, color, object, action, caption). These descriptions provide a structured framework for incorporating new annotations into the Knowledge Graph (KG) as long as they adhere to the specified description criteria. This flexibility ensures that the ontology remains adaptable and capable of accommodating diverse image annotation data.

\clearpage

\section{Evaluation}
\label{eval}

When presented with scenarios relevant to a particular domain, developers should formulate a set of questions that represent user demands and constraints. These questions, known as competency questions (CQs), are fundamental queries about the domain that the ontology aims to address. Aligning scenarios with competency questions ensures that the ontology adequately captures the required knowledge and supports relevant use cases within the organization. Competency questions consist of a set of questions stated in natural language that the ontology must be capable of answering correctly \cite{noy1997state}. Scenario-derived competency questions support the ontology development and evaluation process in two primary ways. Firstly, they enable developers to identify the main elements and their relationships to create the ontology vocabulary (terminology). Secondly, they provide developers with a straightforward means to verify the satisfiability of requirements, either through knowledge retrieval or entailment on its axioms and answers \cite{bezerra}.

Our evaluation protocol consists of several steps aimed at assessing the performance and capabilities of the SituAnnotate system. These steps include the formulation of specialized competency questions, the creation of a toy dataset, the translation of the CQ questions into SPARQL queries, and the execution of these queries over the toy dataset.

\subsection{Competency Questions (CQs) SPARQL Queries}

In the context of the user requirement scenarios, we formulated a set of specialized Competency Questions (CQs). These CQs were designed to reflect the real-world information needs arising from the specific case study and scenarios presented earlier. These questions serve as a valuable tool for assessing the capabilities and performance of the SituAnnotate system in addressing practical use cases. Below, we present the list of CQs derived from our case study and scenarios:

\begin{enumerate}
    \item CQ1: Which countries have been the location of annotation situations, how many annotation situations were located in each country, and which country has been the location for the highest number of annotation situations?
    
    \item CQ2: Between the years 2020 and 2024, in which annotation situations has the image with ID "ARTstract\_14978" been involved?
    
    \item CQ3: What remuneration schemes have been used in annotation situations involving the "ARTstract" dataset?
    
    \item CQ4: What types of entities have been annotated?
    
    \item CQ5: Which images have been annotated using the lexical entry "surfboard," and what role did these annotations serve?
    
    \item CQ6: For the specific situation in which "surfboard" was used to annotate the image with ID "ARTstract\_14978," what contextual factors were associated with the annotation situation?
    
    \item CQ7: Which images have annotations classified under the role of "detected emotion" with an annotation strength exceeding 0.85, and what labels have been assigned to them?
    
    \item CQ8: What concepts type annotations about the image with ID "ARTstract\_14978"?
    
    \item CQ9: For each lexical entry (label) that the image with ID "ARTstract\_14978" was annotated with, who was the Annotator that assigned that label?
    
    \item CQ10: What types of annotations about the image with ID "ARTstract\_14978" were all done by artificial annotators with the "visual transformer" model architecture?
    
    \item CQ11: What are the caption annotations for the image with ID "ARTstract\_14978," and who are the annotators responsible for each caption annotation?
\end{enumerate}

% \clearpage

\subsection{Toy Dataset Creation}

\begin{figure*}[h]
  \includegraphics[width=\textwidth]{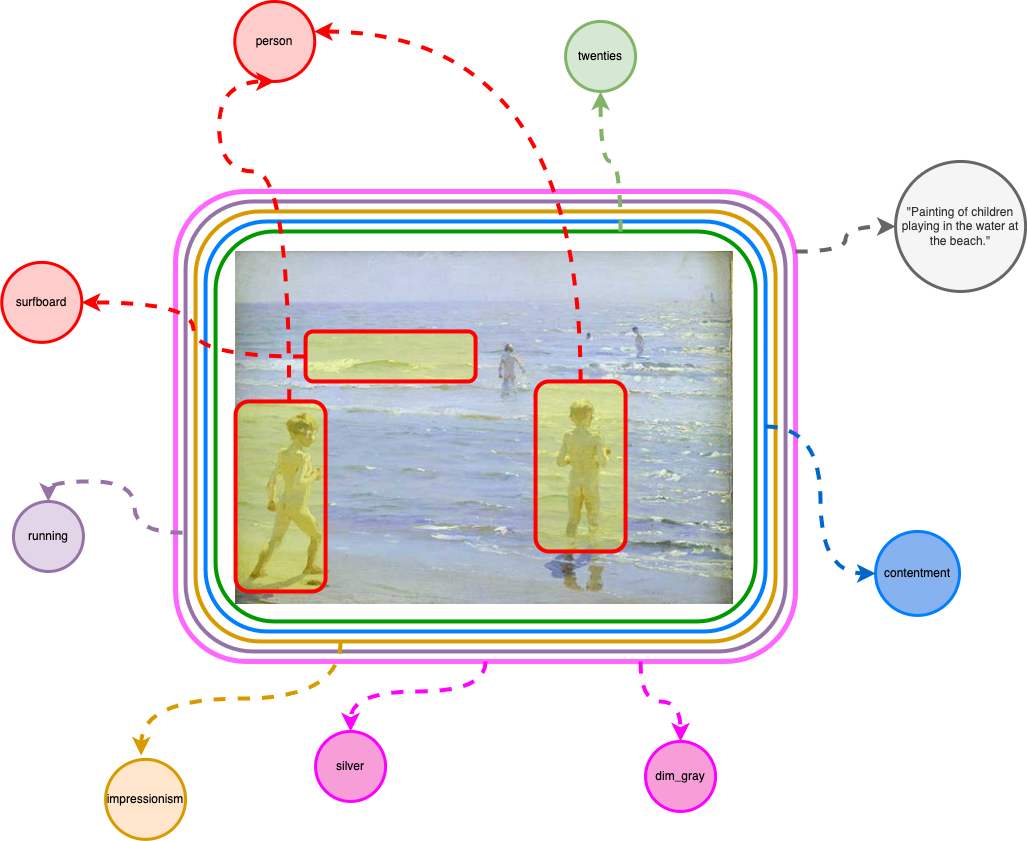}
  \caption{Visualization of the toy dataset, showcasing the image and various labels incorporated into the KG. Each of these labels is hardcoded alongside all of the contextual information about the annotation situation that gave rise to it.}
  \label{fig:toy}
\end{figure*}

To evaluate the capabilities and performance of the SituAnnotate system, we crafted a toy dataset in the form of a Knowledge Graph (KG). This dataset emulates real-world scenarios involving multiple annotation situations for a single image, offering a comprehensive testbed for our system. The toy dataset encompasses a diverse array of annotation types, such as object detection, actions, emotions, art styles, colors, and more, all meticulously generated by distinct artificial annotators. Figure \ref{fig:toy} visually presents the selected image along with the various labels incorporated into the KG, showcasing the rich contextual data captured. 

To formalize the dataset, we employed the SituAnnotate ontology, ensuring the preservation of extensive information pertaining to each annotation situation. This encompassed details like geographical location, temporal specifics, annotated datasets, remuneration structures, detection criteria, and detailed annotator profiles. This rich contextual data not only enhances the semantic content of the dataset but also enables structured representation for diverse analytical purposes.

\subsubsection{Image Data Knowledge Graph}: This RDF file, available \href{https://github.com/delfimpandiani/situAnnotate/blob/main/tests/toy_dataset/images_kg.ttl}{here}, provides comprehensive data for a set of annotations related to a specific image (\textit{:ARTstract\_14978}). Each annotation within the dataset is associated with the annotation situation in which it took place. These annotations span various dimensions, including actions, age groups, artistic styles, colors, emotions, human presence, image captions, and objects, all linked to relevant ConceptNet concepts. Moreover, each annotation is enriched with an annotation strength value, reflecting its confidence or relevance. 
% Below is an example of the triples each annotation has:
% \begin{verbatim}
% :14978_ARTstract_as_2023_06_26 a :ArtStyleAnnotation ;
%     :aboutAnnotatedEntity :ARTstract_14978 ;
%     :annotationWithLexicalEntry :le_Impressionism ;
%     :hasAnnotationStrength 0.6149182915687561 ;
%     :isAnnotationInvolvedInSituation :ARTstract_as_2023_06_26 ;
%     :isClassifiedBy :detected_art_style ;
%     :typedByConcept <http://etna.istc.cnr.it/framester2/conceptnet/5.7.0/c/en/impressionism> .
% \end{verbatim}

\subsubsection{Annotation Situations Knowledge Graph (KG)}: This RDF file, accessible \href{https://github.com/delfimpandiani/situAnnotate/blob/main/tests/toy_dataset/situations_kg.ttl}{here}, contains detailed representations of the annotation situations themselves, including details about geographical locations, dates, annotators, and more. Notably, this KG incorporates further information about the artificial annotators used for generating annotations. These annotators are associated with specific model architectures and datasets. 

% Below is an example of the triples a single AnnotationSituation may have. 
% \begin{verbatim}
% :ARTstract_as_2023_06_26 a :ArtStyleAnnotationSituation ;
%     :involvesAnnotatedEntity :ARTstract_14978 ;
%     :atPlace :Italy ;
%     :hasDetectionThreshold "top_one" ;
%     :involvesAnnotator :oschamp_vit-artworkclassifier ;
%     :involvesDataset :ARTstract ;
%     :onDate "2023-06-26"^^xsd:date ;
%     :satisfies :as_detection_desc .

% :as_detection_desc a ns1:ImageAnnotationDescription ;
%     rdfs:comment "Art style detections are annotation situations 
%         in which annotations play the role of detected_art_style, 
%         assigned by an Annotator according to a certain detection 
%         threshold or heuristic"^^xsd:string ;
%     :defines :detected_art_style .

% :oschamp_vit-artworkclassifier a :ArtificialAnnotator ;
%     :hasModelArchitecture :visual_transformer ;
%     :pretrainedOnDataset :artbench-10 .
% \end{verbatim}

\subsection{Translation into and Execution of SPARQL Queries}

These CQs were subsequently translated into SPARQL queries, creating a formal means to retrieve specific information from the toy dataset. To evaluate the performance and effectiveness of the SituAnnotate system, we executed these SPARQL queries over the toy dataset. For executing the SPARQL queries, we used Ontotext GraphDB,\footnote{\url{https://graphdb.ontotext.com/documentation/10.0/index.html}} a highly efficient and robust graph database with RDF and SPARQL support. We ran GraphDB in a Docker container, as provided on Github.\footnote{\url{https://github.com/Ontotext-AD/graphdb-docker}} This platform facilitated the execution of SPARQL queries and retrieval of structured data in accordance with the SituAnnotate ontology. To provide a concise overview, Table \ref{tab:sparql} summarizes the Competency Questions (CQs) along with their corresponding SPARQL queries and whether they were successfully executed ("Pass" status) in evaluating the SituAnnotate system's performance.

\subsubsection{Results} All 11 competency question verification tests were successfully passed, with the expected outcomes matching the actual results. Comprehensive details regarding the results can be accessed in our SituAnnotate GitHub repository. \footnote{\url{https://github.com/delfimpandiani/situAnnotate/blob/main/tests/competency_question_verification/Results.md}} This repository provides in-depth insights into the query outcomes, presenting the retrieved information relevant to each specialized competency question. We invite readers to visit our online repository for a thorough comprehension of our evaluation process and results, facilitating further analysis as needed.

\subsection{Human-Understandable Explanations: SPARQL to Natural Language}

To further enhance SituAnnotate's utility, we developed a method to translate SPARQL query results into human-readable explanations, facilitating comprehension and transparency within annotation contexts.

\paragraph*{Development of SPARQL Query}
We crafted a specialized SPARQL query capable of retrieving comprehensive contextual information for a given entity-label pair. This query is designed to extract relevant situational details such as annotator demographics, model architecture, and dataset information.

\paragraph{Translation with Python Script}
Subsequently, we developed a Python script to automatically translate the SPARQL query output into a human-readable narrative. This script parses the SPARQL query results and generates a coherent explanation of the annotation context, presenting it in a format easily understandable by individuals with diverse knowledge backgrounds.

\paragraph{Practical Application and Demonstration}
To showcase the practical application of this approach, we selected an image from our toy dataset and executed the SPARQL query along with the Python script. The resulting human-readable explanation provided insightful context surrounding the annotation, including details about the annotator, model architecture, and dataset used for training.

\begin{figure*}[h]
\includegraphics[width=\textwidth]{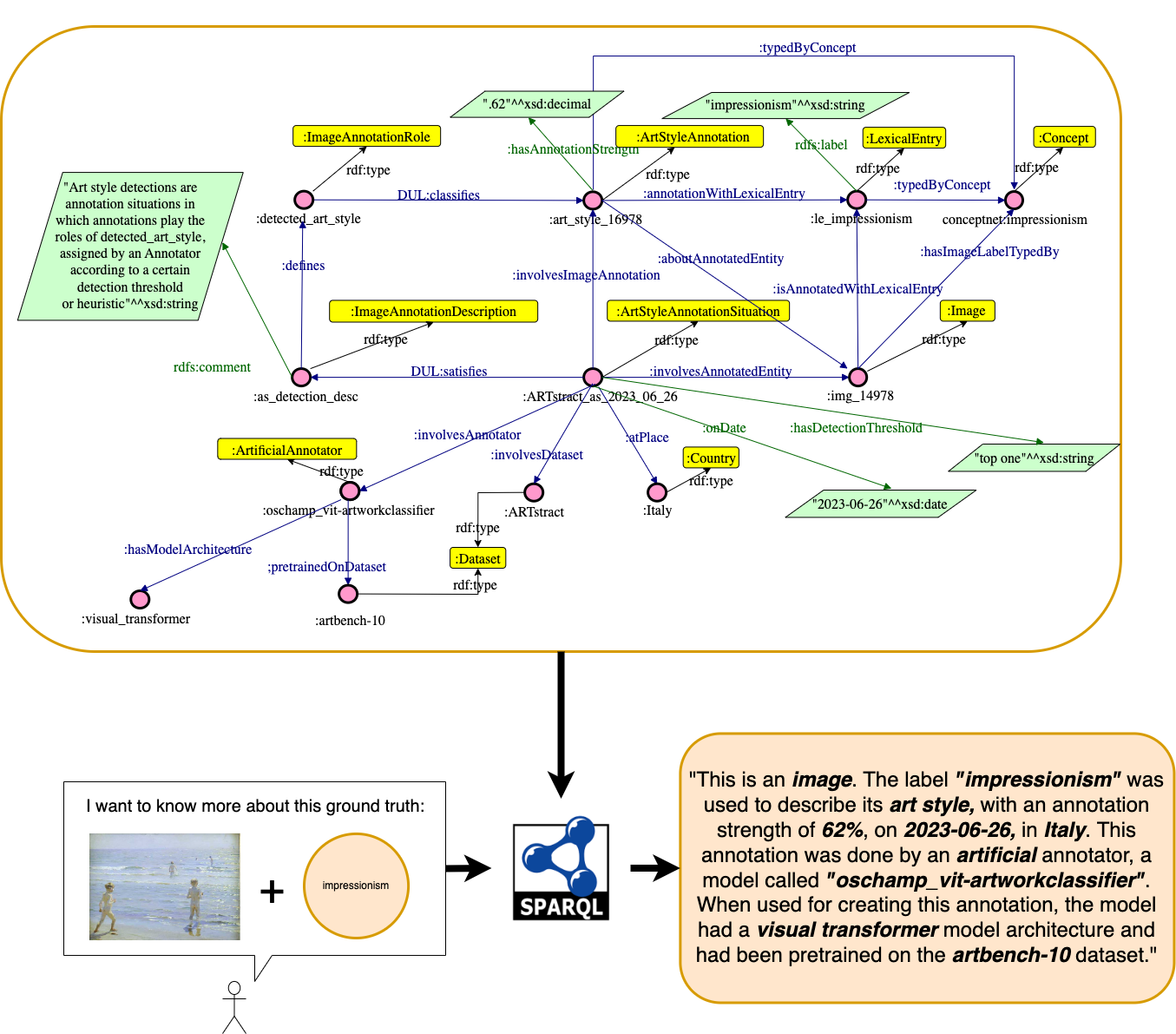}
\caption{Illustration of an exemplary usage of the ImageAnnotationSituation specialization, depicting the formal representation of the contextual situation in which the label "impressionism" was assigned to a specific image. Users can employ a single SPARQL query and a small Python script to obtain fully human-understandable statements situating the ground truth label assigned to an entity, as demonstrated in this case with an image.}
\label{fig:user_query}
\end{figure*}

\paragraph{Results and Implications}
Figure \ref{fig:user_query} illustrates an exemplary human-readable explanation generated using this method for an image labeled as "impressionism." This practical demonstration underscores SituAnnotate's ability to provide insightful and human-understandable explanations of annotations, thereby enhancing transparency and interpretability in AI systems.

\paragraph{Availability}
The Python script for executing and translating the SPARQL query into natural language is openly accessible online, facilitating its adoption and adaptation by the research community.\footnote{\url{github.com/delfimpandiani/situAnnotate/tests/}}

\section{Discussion}
\label{discussion}

The positive evaluation results highlight the robustness and power of SituAnnotate in formally representing information. These results indicate that SituAnnotate excels in several key aspects, providing significant advantages over traditional annotation methods.

\paragraph{\textbf{Contextual Precision}} SituAnnotate provides a highly nuanced and context-aware representation of annotations. By connecting annotations not only to the described entity but also to the broader situational context and the annotator involved, it enables a richer understanding of the circumstances in which annotations are made. This contextual precision is often lacking in traditional annotation approaches that focus solely on labels or strings.

\paragraph{\textbf{Semantic Enrichment and Expressiveness}} Unlike traditional annotation methods that often involve fixed annotation tasks and roles, SituAnnotate offers the flexibility of associating the same entity with multiple labels through various "AnnotationRoles." This semantic depth significantly enhances the ontology's expressiveness and precision. Annotators can provide richer and more detailed information about the same entity, enabling a more comprehensive understanding of the annotated data. This flexibility is particularly valuable when dealing with complex or multifaceted annotations. SituAnnotate can even formally represent cases in which the same entity, e.g. an image, is associated with the same label, e.g. "happiness," through various annotations fulfilling different AnnotationRoles, such as detected emotion or detected abstract concept. This semantic depth significantly enhances the ontology's expressiveness and precision

\paragraph{\textbf{Flexibility in Annotator Representation}} SituAnnotate is adaptable to various annotator types, including both artificial and human annotators. This adaptability addresses privacy concerns by allowing the aggregation of demographic data when needed. In today's diverse annotation landscape, which includes computer vision models, individual human annotators, and annotator communities, SituAnnotate ensures that all these entities can be formally represented. This reflects the multifaceted nature of modern annotation processes and supports inclusive and diverse annotation practices.

\paragraph{\textbf{Automated Reasoning and Data-Driven Decision-Making}} 
SituAnnotate's structured representation of annotation situations facilitates automated reasoning through SPARQL queries and semantic technologies. Machines can infer relationships, make connections, and retrieve information efficiently, streamlining the annotation understanding process. This automation not only saves time but also supports data-driven decision-making. Users can leverage SituAnnotate to extract valuable insights and patterns from annotated data, enabling evidence-based decisions and enhancing the utility of labeled datasets.

\paragraph{\textbf{Enhanced and User-Friendly Human Understanding}} SituAnnotate, despite its machine-readable foundation, offers a user-friendly ontological framework that remains accessible to researchers, domain experts, and annotators alike. This approach ensures that the ontology isn't confined to machines but serves as a valuable resource for human understanding. Moreover, the integration of SPARQL queries and Python scripts empowers users to effortlessly access and interpret situational knowledge tied to specific annotations. This user-friendly feature enhances transparency and facilitates comprehension, making SituAnnotate a versatile resource catering to both machine-driven AI technologies and human expertise. This symbiotic relationship fosters a deeper synergy between AI capabilities and human insights, emphasizing the ontology's significance in bridging the gap between technology and human cognition.

\paragraph{\textbf{Comparing Annotation Situations for Enhanced Understanding}} SituAnnotate's ability to represent multiple labels and annotation situations related to the same annotated object provides users with a powerful tool for enhanced understanding. Through SPARQL queries, users can retrieve all AnnotationSituations for an object, enabling detailed comparisons of potentially conflicting interpretations. This feature enhances the understanding of diverse annotation contexts and their implications, supporting better decision-making and data analysis.

\paragraph{\textbf{Mitigating Bias and Enhancing Ethical AI}} SituAnnotate serves as a robust tool in the battle against bias through its context-aware data annotation capabilities. Annotators can furnish essential details about data sources, annotator demographics, and the rationale behind labeling decisions. This wealth of contextual information empowers AI developers to scrutinize and rectify any latent biases when examining annotated data. By doing so, SituAnnotate champions transparency and fairness throughout the data annotation process. An example of this usage is found in \cite{martinez2024stitching}. SituAnnotate's ontology was employed to model the situated assignment of annotation labels to images, considering various levels of abstraction and annotation roles such as objects, actions, and emotions. This allowed for a detailed examination of the annotation context, facilitating the tracking of annotations by different annotators and at diverse levels of abstraction. Consequently, a knowledge graph was generated, comprising comprehensive triples that encapsulated the diverse aspects of annotation situations concerning images. This laid the foundation for a thorough analysis of biases ingrained within the dataset. By utilizing SPARQL queries on the knowledge graph, posthoc interpretability analyses were conducted, offering insights into the biases present in the annotated dataset. For instance, researchers could distinguish annotations generated with visual transformers from those produced by other methods. This distinction enabled a nuanced analysis of biases associated with visual transformer-generated annotations compared to other methods. Through this approach, patterns and correlations within the dataset were identified, further validating the presence of perceptual topology biases and guiding strategies for bias mitigation. An example of such analysis is depicted in Figure 7 of that manuscript.

\subsection{Limitations}

Despite its promising capabilities, SituAnnotate does have some limitations and challenges:

\begin{enumerate}
    \item \textbf{Knowledge Representation Overhead:} While SituAnnotate offers enhanced contextual knowledge representation, this also introduces an overhead in terms of ontology creation, maintenance, and population. It may require substantial time and effort to initially set up and continuously update.

    \item \textbf{Capturing Human Subjectivity and Cultural Nuances:} One notable challenge lies in the complexity of capturing the full scope of contextual factors that affect human subjectivity and the diverse cultural nuances that can influence annotations. While SituAnnotate offers a structured framework, it does not fully capture the richness of human interpretation.

    \item \textbf{Scalability Concerns:} SituAnnotate's scalability may be a concern when applied to massive datasets, where managing a vast number of annotation situations and annotators can become unwieldy. Optimizing the ontology for large-scale applications is an ongoing challenge. This is also because the use of SPARQL queries and scripts to retrieve human-understandable explanations can be resource-intensive.

    \item \textbf{Privacy Mechanisms:} SituAnnotate's ability to address privacy concerns may require further refinement to provide more robust mechanisms for data anonymization and aggregation. Ensuring the privacy and confidentiality of sensitive data is crucial.

    \item \textbf{Cost and Time Implications:} An important consideration is the potential increase in cost and time associated with implementing SituAnnotate. While the framework provides more comprehensive information for each annotated object compared to standard schemes, it does require additional resources for annotation tasks, including time, memory, and potentially increased payments to annotators. We acknowledge the need for a comparative analysis with state-of-the-art annotation frameworks to provide insights into the cost and time implications. Such an analysis would contribute to a better understanding of the trade-offs involved and help inform decision-making in adopting SituAnnotate.
    
\end{enumerate}

These limitations should be considered when implementing SituAnnotate in real-world scenarios, and ongoing research and development efforts may help mitigate some of these challenges.

\subsection{Further Directions}

As SituAnnotate continues to evolve, there are several avenues for further research and development:

\begin{enumerate}
    \item \textbf{Usability Improvements:} Prioritize creating user-friendly tools and interfaces that simplify the process of integrating SituAnnotate into annotation workflows. Consider developing user-friendly graphical user interfaces (GUIs) for creating and querying annotations, enhancing accessibility for a broader user base.

    \item \textbf{Scalability:} Investigate methods to enhance SituAnnotate's scalability, particularly when dealing with large datasets. This may involve optimizing SPARQL queries or exploring distributed computing solutions to handle increasing volumes of data efficiently.

    \item \textbf{Enhanced Automation:} Continue to advance automation tools for generating human-readable explanations from the ontology. Explore Natural Language Processing (NLP) techniques to produce more coherent and concise explanations, reducing the need for manual intervention.

    \item \textbf{Interoperability:} Ensure that SituAnnotate remains compatible with other ontologies and standards in the data annotation and semantic web domain. Seamless integration with existing systems is essential for broader adoption.

    \item \textbf{Community Involvement:} Foster collaboration and engagement within the research community to refine and expand SituAnnotate. An active user and developer community can drive further innovation and adoption. Additionally, seek collaboration with the global research community to address cultural biases and diversify the ontology's applicability.

    \item \textbf{Ethical Considerations:} Delve into the ethical implications of SituAnnotate's real-world applications, particularly concerning privacy, bias, and transparency. Develop comprehensive guidelines and best practices for ethical annotation processes, promoting responsible AI development.

    \item \textbf{Scenarios and Use Cases:} Continue to develop and document a diverse set of real-world scenarios and use cases where SituAnnotate has demonstrated its practical value. Providing concrete examples can help potential users grasp its applicability better.

    \item \textbf{Integration with AI Systems:} Explore seamless integration possibilities of SituAnnotate with AI systems, particularly in domains like computer vision, natural language processing, and knowledge graphs. Incorporating advanced techniques for handling multi-modal data, including text, images, and videos, can broaden its applicability.

    \item \textbf{AI Ethics and Fairness:} Investigate how SituAnnotate can be integrated with emerging AI ethics and fairness frameworks. Contributing to more responsible and equitable AI development aligns with the growing importance of ethical considerations in the field.

\end{enumerate}

These directions encompass technical, usability, ethical, and collaborative aspects, ensuring that SituAnnotate remains a dynamic and relevant tool in the evolving landscape of data annotation and semantic web technologies.

\section{Conclusion}
\label{conclusion}

In conclusion, the SituAnnotate ontology provides a robust and context-aware framework for situating ground truths, i.e., representing annotations within the contextual situations from which they arise. Aligned with the Dolce Ultra Light ontology, it ensures consistency and interoperability, while its expressive relationships and semantic depth enhance annotation context understanding. Researchers and practitioners can use SituAnnotate to model, analyze, and interpret annotations in a structured and standardized way, making it a valuable contribution to data annotation and knowledge representation. SituAnnotate overcomes traditional annotation method limitations, benefiting both human annotators and automated processes with a structured, machine-readable format that remains human-readable. Its SPARQL query support enables efficient data retrieval and analysis, bridging the gap between structured data and human comprehension, enhancing annotation efficiency and accuracy, and promoting transparency and ethical considerations in data annotation—a crucial step for responsible AI development. Ultimately, SituAnnotate's contextual annotations enhance AI decision-making, aiding models in adapting to real-world scenarios and advancing ethical AI implementation.

\begin{landscape}
\centering
\footnotesize
% \caption{SituAnnotate Competency Questions and Corresponding SPARQL Queries}
\label{tab:sidewaystable}
\begin{xtabular}{|p{0.5cm}|p{3.5cm}|p{10cm}|p{0.5cm}|}
\hline
\textbf{CQ} & \textbf{Competency Question} & \textbf{SPARQL Query} & \textbf{Pass} \\
% \endfirsthead
\hline
% __________________________________________________________
CQ1 & Which countries have been the location of annotation situations, how many annotation situations were located in each country, and which country has been the location for the highest number of annotation situations? & \begin{lstlisting}[style=sparql]
SELECT ?Country (COUNT(?AnnotationSituation) AS ?count)
WHERE {
    ?AnnotationSituation :atPlace ?Country .
}
GROUP BY ?Country
ORDER BY DESC(?count)
\end{lstlisting} & Y \\
\hline
% __________________________________________________________
CQ2 & Between the years 2020 and 2024, in which annotation situations has the image with ID \textit{ARTstract\_14978} been involved? & \begin{lstlisting}[style=sparql]
SELECT ?AnnotationSituation ?Date
WHERE {
    :ARTstract_14978 :isInvolvedInAnnotationSituation ?AnnotationSituation .
    ?AnnotationSituation :onDate ?Date .
    FILTER(YEAR(?date) >= 2020 && YEAR(?date) <= 2024)
}
\end{lstlisting} & Y \\
\hline
% __________________________________________________________
CQ3 & What remuneration schemes have been used in annotation situations involving the \textit{ARTstract} dataset? & \begin{lstlisting}[style=sparql]
SELECT ?RemunerationScheme 
WHERE {
    ?AnnotationSituation rdf:type :AnnotationSituation ;
    :involvesDataset :ARTstract .
    ?AnnotationSituation :involvesRemunerationScheme ?RemunerationScheme .
}
\end{lstlisting} & Y \\
\hline
% __________________________________________________________
CQ4 & What types of entities have been annotated? & \begin{lstlisting}[style=sparql]
SELECT DISTINCT ?EntityType 
WHERE {
    ?Annotation :aboutAnnotatedEntity ?Entity .
    ?Entity a ?EntityType .
}
\end{lstlisting} & Y \\
\hline
\hline
% \endfoot
% __________________________________________________________
CQ5 & Which images have been annotated using the lexical entry "surfboard," and what role did these annotations serve? & \begin{lstlisting}[style=sparql]
SELECT ?Image ?annotationRole 
WHERE {
    ?Annotation :aboutAnnotatedEntity ?Image .
    ?Annotation :annotationWithLexicalEntry :le_surfboard .
    ?Annotation :isClassifiedBy ?AnnotationRole .
}
\end{lstlisting} & Y \\
\hline
% __________________________________________________________
CQ6 & For the specific situation in which "surfboard" was used to annotate the image with ID \textit{ARTstract\_14978}, what contextual factors were associated with the annotation situation? & \begin{lstlisting}[style=sparql]
SELECT ?Country ?Date ?Dataset ?RemunerationScheme ?DetectionThreshold ?Annotator ?PretrainDataset ?ModelArchitecture 
WHERE {
  ?Annotation :aboutAnnotatedEntity :ARTstract_14978 .
  ?Annotation :annotationWithLexicalEntry :le_surfboard .
  ?AnnotationSituation :involvesAnnotation ?Annotation .
  OPTIONAL {
      ?AnnotationSituation :atPlace ?Country .
      ?AnnotationSituation :onDate ?Date .
      ?AnnotationSituation :involvesDataset ?Dataset .
      ?AnnotationSituation :hasDetectionThreshold ?DetectionThreshold .
      ?AnnotationSituation :involvesAnnotator ?Annotator .
      ?Annotator :pretrainedOnDataset ?PretrainDataset .
      ?Annotator :hasModelArchitecture ?ModelArchitecture .
      ?AnnotationSituation :involvesRemunerationScheme ?RemunerationScheme .
  }
}
\end{lstlisting} & Y \\
\hline
% __________________________________________________________
CQ7 & Which images have annotations classified under the role of "detected emotion" with an annotation strength exceeding 0.85, and what labels have been assigned to them? & \begin{lstlisting}[style=sparql]
SELECT ?Image ?Label
WHERE {
  ?Image a :Image .
  ?Annotation :aboutAnnotatedEntity ?Image ;
              :isClassifiedBy :detected_emotion ;
              :hasAnnotationStrength ?AnnotationStrength ;
              :annotationWithLexicalEntry ?LE .
  ?LE rdfs:label ?Label .
  FILTER (?AnnotationStrength > 0.85)
}
\end{lstlisting} & Y \\
\hline
% __________________________________________________________
CQ8 & What concepts type annotations about the image with ID \textit{ARTstract\_14978}? & \begin{lstlisting}[style=sparql]
SELECT ?Concept ?AnnotationRole ?AnnotationStrength
WHERE {
    ?Annotation :aboutAnnotatedEntity :ARTstract_14978 .
    ?Annotation :isClassifiedBy ?AnnotationRole .
    ?Annotation :hasAnnotationStrength ?AnnotationStrength .
    ?Annotation :typedByConcept ?Concept .
}
\end{lstlisting} & Y \\
\hline
% __________________________________________________________
CQ9 & For each lexical entry (label) that the image with ID \textit{ARTstract\_14978} was annotated with, who was the Annotator that assigned that label? & \begin{lstlisting}[style=sparql]
SELECT ?string ?Annotator 
WHERE {
    :ARTstract_14978 :isInvolvedInAnnotationSituation ?AnnotationSituation .
    ?AnnotationSituation :involvesAnnotation ?Annotation .
    ?AnnotationSituation :involvesAnnotator ?Annotator .
    ?Annotation :aboutAnnotatedEntity :ARTstract_14978.
    ?Annotation :annotationWithLexicalEntry ?LexicalEntry .
    ?LexicalEntry rdfs:label ?string .
}
\end{lstlisting} & Y \\
\hline
% __________________________________________________________
CQ10 & What types of annotations about the image with ID \textit{ARTstract\_14978} were all done by artificial annotators with the \textit{visual transformer} model architecture? & \begin{lstlisting}[style=sparql]
SELECT ?AnnotationClass ?Annotator ?Dataset
WHERE {
    ?AnnotationSituation :involvesAnnotation ?Annotation .
    ?AnnotationSituation :involvesAnnotator ?Annotator .
    ?Annotator :hasModelArchitecture :visual_transformer .
    ?Annotator :pretrainedOnDataset ?Dataset .
    ?Annotation :aboutAnnotatedEntity :ARTstract_14978 .
    ?Annotation a ?AnnotationClass .
    FILTER NOT EXISTS {
        ?subClass rdfs:subClassOf ?AnnotationClass .
        ?Annotation rdf:type ?subClass.
        FILTER (?subClass != ?AnnotationClass)
    }
}
\end{lstlisting} & Y \\
\hline
% __________________________________________________________
CQ11 & What are the caption annotations for the image with ID \textit{ARTstract\_14978}, and who are the annotators responsible for each caption annotation? & \begin{lstlisting}[style=sparql]
SELECT ?Caption ?Annotator
WHERE {
  ?Annotation :aboutAnnotatedEntity :ARTstract_14978 .  
  ?Annotation a :ImageCaptionAnnotation .
  ?Annotation rdfs:comment ?Caption .
  ?AnnotationSituation :involvesAnnotation ?Annotation .
  ?AnnotationSituation :involvesAnnotator ?Annotator .
}
\end{lstlisting} & Y \\
\hline
\end{xtabular}
\label{tab:sparql}
\end{landscape}

\bibliographystyle{ACM-Reference-Format}
\bibliography{main}
\end{document}